# Real-time Cardiovascular MR with Spatio-temporal Artifact Suppression using Deep Learning - Proof of Concept in Congenital Heart Disease

Andreas Hauptmann (PhD)[1],
Simon Arridge (PhD)[1],
Felix Lucka (PhD)[1,2],
Vivek Muthurangu (MD)[3*],
Jennifer A. Steeden (PhD)[3*]

*Authors V. Muthurangu and J. Steeden contributed equally to this work*

1. Department of Computer Science, University College London, London. WC1E 6BT. United Kingdom
2. Computational Imaging, Centrum Wiskunde & Informatica (CWI), Science Park 123, 1098 XG Amsterdam
3. UCL Centre for Cardiovascular Imaging, University College London, London. WC1N 1EH. United Kingdom

**Corresponding Author:** Vivek Muthurangu

UCL Centre for Cardiovascular Imaging, Institute of Cardiovascular Science

30 Guildford Street, London. WC1N 1EH

v.muthurangu@ucl.ac.uk

Tel: +44 (0)207 762 6834

Fax: +44 (0)207 813 8263

# ABSTRACT

## Purpose

Real-time assessment of ventricular volumes requires high acceleration factors. Residual convolutional neural networks (CNN) have shown potential for removing artifacts caused by data undersampling. In this study, we investigated the ability of CNN's to reconstruct highly accelerated radial real-time data in patients with congenital heart disease (CHD).

## Methods

A 3D (2D plus time) CNN architecture was developed, and trained using synthetic training data created from previously acquired breath hold cine images from 250 CHD patients. The trained CNN was then used to reconstruct actual real-time, tiny Golden Angle (tGA) radial SSFP data (13x undersampled) acquired in 10 new patients with CHD. The same real-time data was also reconstructed with compressed sensing (CS) to compare image quality and reconstruction time. Ventricular volume measurements made both using the CNN and CS reconstructed images were compared to reference standard breath hold data.

## Results

It was feasible to train a CNN to remove artifact from highly undersampled radial real-time data. The overall reconstruction time with the CNN (including creation of aliased images) was shown to be >5x faster than the CS reconstruction. In addition, the image quality and accuracy of biventricular volumes measured from the CNN reconstructed images were superior to the CS reconstructions.

## Conclusion

This paper has demonstrated the potential for the use of a CNN for reconstruction of real-time radial data, within the clinical setting. Clinical measures of ventricular volumes using real-time data with CNN reconstruction are not statistically significantly different from gold-standard, cardiac gated, breath-hold techniques.

# INTRODUCTION

Cardiovascular magnetic resonance (CMR) is the reference standard method of measuring ventricular volumes in patients with congenital heart disease (CHD). Conventionally, breath-hold (BH), multi-slice, short-axis, cardiac gated, balanced steady state free precession (bSSFP) cine imaging is used to acquire this data (1). Unfortunately, many patients find breath-holding difficult and a rapid free breathing alternative is desirable.

Real-time imaging is a possible solution, but relies on high levels of data undersampling to ensure adequate spatial and temporal resolution. This means that sophisticated reconstruction techniques are necessary to produce artifact-free images. One example is compressed sensing (CS), which is increasingly used to reconstruct real-time CMR images (2-4). However, there are drawbacks to CS including: i) Computationally intensive and time-consuming reconstruction, and ii) Challenging optimisation of reconstruction parameters, which can result in unnatural looking images. Thus, alternative reconstruction approaches may be useful.

The removal of aliases in images from undersampled MR data, can be thought of as a de-noising problem. This is particularly true if the sampling pattern produces incoherent artifacts, as is the case with tiny golden angle (tGA) radial spokes (5). Recently, convolutional neural networks (CNN) have been shown to be well suited to de-noising and artifact removal problems (6-8). Consequently, it may be possible to use CNN's to remove aliasing from undersampled real-time MR data. This approach is dependent on large amounts of high quality training data and one potential source is previously acquired, retrospectively cardiac gated BH-bSSFP cine imaging.

In this study, 'synthetic' training data was created from a library of BH-bSSFP cine images and used to train a CNN to map between aliased and artifact-free images (deep artifact suppression). The trained CNN was then used to remove aliases from actual undersampled real-time MR data acquired in prospectively scanned patients with CHD. The aims of this study were to: i) Investigate the effect of different radial sampling patterns on the effectiveness of CNN based reconstruction, ii) Acquire actual real-time data in patients with CHD, using a tGA radial sampling pattern, iii) Compare reconstruction speed, image quality and accuracy of ventricular volumes calculated from CNN reconstruction, compared to a current state-of-the-art CS algorithm.

# METHODS

## Deep Artifact Suppression Algorithm – General concept

The deep artifact suppression algorithm was based on a modified residual U-Net architecture (9,10). A residual U-Net is a multi-scale CNN where images are sequentially down-sampled and then up-sampled to produce an updated and clean version of the artifact-contaminated input image. This particular structure is especially suitable for tomographic problems (such as MR imaging), where the normal operator (projection followed by back-projection) is a type of convolution (10). The effectiveness of this approach has been shown in several studies for 2D tomographic problems (11,12) and in 2.5D (6), where the temporal information is processed as 2D channels. In our study, the input was a 3D data set (2D image through time). A 3D structure was used to enforce temporal consistency between frames and prevented the flickering artifact present if a 2D slice-by-slice approach is used (see Supporting Information Video S1).

The residual U-Net was trained with synthetic training data consisting of paired artifact-free 'ground truth' magnitude images and corresponding undersampled, artifact-contaminated images. The trained network was then tested with synthetic test data to quantify performance. Finally, the same trained network was used to reconstruct actual real-time data acquired in patients with CHD. These steps are fully explained below. The use of retrospectively collected training and test data, as well as collection of prospective real-time data was approved by the local research ethics committee, and written consent was obtained from all subjects/guardians (Ref: 06/Q0508/124).

## Preparation of Synthetic Training Data

The synthetic training data was created from 2D short axis stacks of Cartesian, retrospectively cardiac gated, breath hold, multi-slice, short-axis, bSSFP cine images (BH-bSSFP). The imaging parameters were: matrix ~240x204, FOV ~320x265 mm, phase partial-Fourier 6/8, spatial resolution ~1.3x1.3 mm, slice thickness ~10mm, temporal resolution ~32 ms, reconstructed cardiac phases 40. Training data was collected from 250 previously scanned children and adults with paediatric heart disease or CHD (mean age: 22.3±12.6 years, male: 140). No data from patients with single ventricles or images with breathing or arrhythmia artifacts were included in the synthetic training data. Full details of patient diagnoses are supplied in Supporting Information S2. On average, each patient had ~9 2D slices, resulting in 2276 paired 3D (2D plus time) data sets for training.

The first step in creating training data was to convert the retrospectively cardiac gated data into synthetic real-time data that mirrored the acquisition parameters of the real-time sequence (as stated below). To provide similar spatial resolution, images were linearly resampled onto a 192x192 matrix, reducing the spatial resolution from 1.3 mm to 1.7 mm. This was followed by linear resampling in time onto points that were 36.4 ms apart (temporal resolution of the real-time sequence). The total number of temporal frames equalled the floor of the R-R wave divided by 36.4 ms. The resulting data formed the 'ground truth' images necessary for training.

To create undersampled, artifact-contaminated training data, the synthetic real-time data was Fourier transformed and gridded onto a 13x undersampled radial trajectory (see below). The undersampled data was then regridded and inverse Fourier transformed back into image space. Both the ground truth and artifact-contaminated images were cropped to a 128x128 matrix to constrain the learning problem to the anatomy of interest (heart). The ground truth and artifact-contaminated images were also linearly interpolated through time, to give 20 frames, as all inputs to the network had to be of the same dimensions. Finally, each 3D data set was also normalized to have signal intensities in the range [0, 1]. All processing required for creation of the synthetic training data was performed in MATLAB

(MATLAB 2016b, The MathWorks, Inc., Natick, Massachusetts, United States). A flow diagram of the process is included in Supporting Information Figure S3.

## Assessing the Effect of Sampling Strategy

Radial sampling was chosen for this study, as it is known to produce less pronounced undersampling artifacts compared to Cartesian acquisitions (13). We further hypothesised that a continuously rotating tiny golden angle (~23.63°) approach may be optimal for a residual U-Net reconstruction due to the noise-like characteristics of the aliases. Although tiny golden angles spacing has been shown to reduce eddy current effects and optimize image quality (13), the effect of specific sampling patterns on deep artifact suppression had not been investigated. Therefore, we compared the continuously rotating tiny golden angle sampling ($tGA_{rot}$) to three other sampling patterns (see Figure 1): i) A tiny golden angle scheme with no rotation between frames ($tGA_{no\_rot}$), ii) A regular angle scheme with regular rotation between frames ($REG_{rot}$), and iii) A regular angle scheme with no rotation between frames ($REG_{no\_rot}$). Figure 2 shows both the sampling pattern and their corresponding artifacts.

**Figure 1** *The four radial sampling patterns tested in this study. Fully sampled k-space requires 182 uniformly spaced radial spokes. An acceleration factor of 13 was used, resulting in 14 radial spokes acquired per frame, regardless of sampling pattern.* $REG_{no\_rot}$ – regular angle scheme (equal angular spacing of ~12.9° between each profile) with no rotation between frames. $REG_{rot}$ – regular angle scheme with regular rotation (of ~0.99°) between frames, so that 13 consecutive frames constitutes a fully sampled k-space. $tGA_{no\_rot}$ – tiny golden angle scheme with no rotation between frames. $tGA_{rot}$ – continuously rotating tiny golden angle sampling.

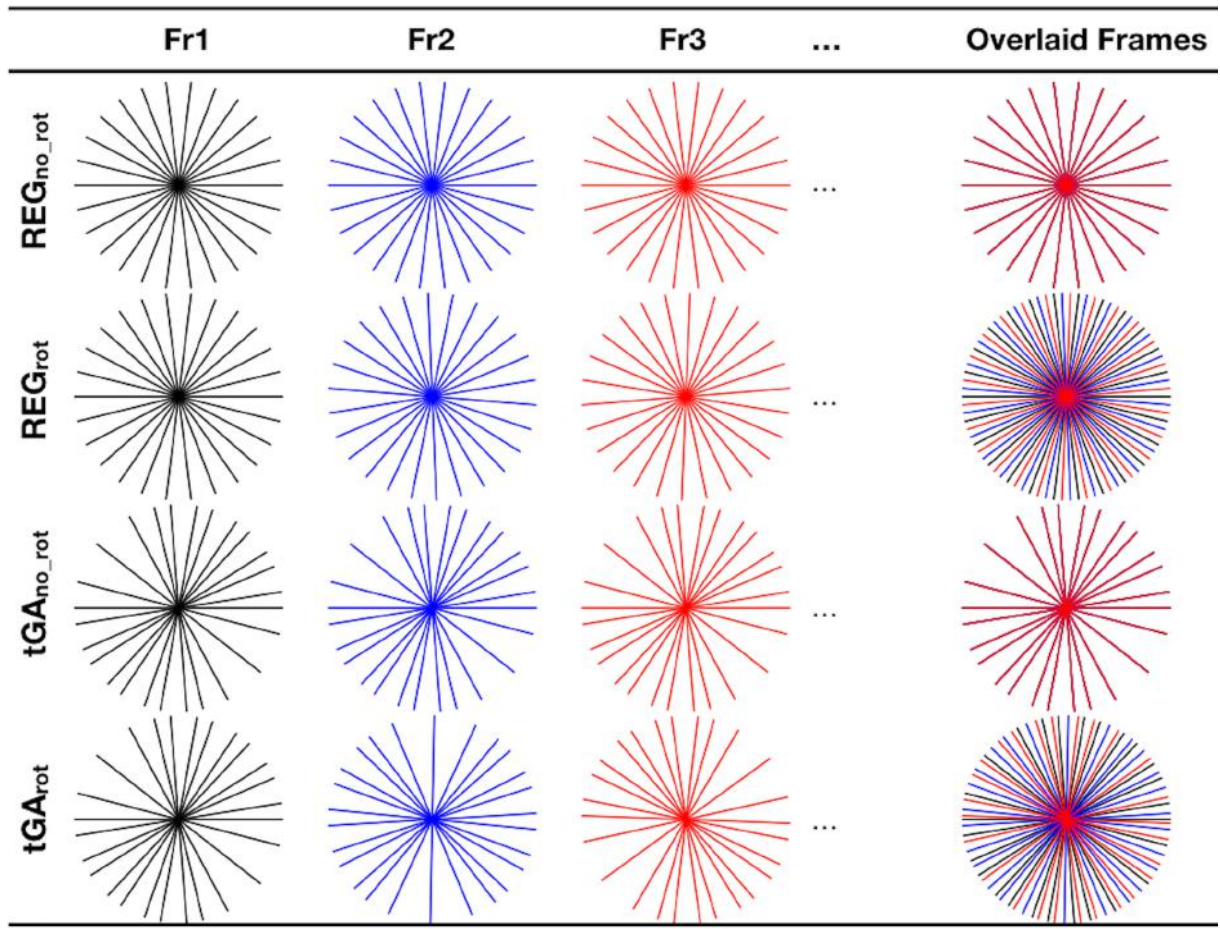

**Figure 2** *Typical artifact produced from the four radial sampling patterns tested in this study. Left, shows a single frame from the cine. Right, shows one line through the ventricles (shown in red) through time. $REG_{no\_rot}$ produces regular aliases with no variation through time. $REG_{rot}$ produces regular aliases in space that change position over time. $tGA_{no\_rot}$ produces spatially incoherent aliases that do not vary through time. $tGA_{rot}$ produces spatially incoherent aliases that vary through time.*

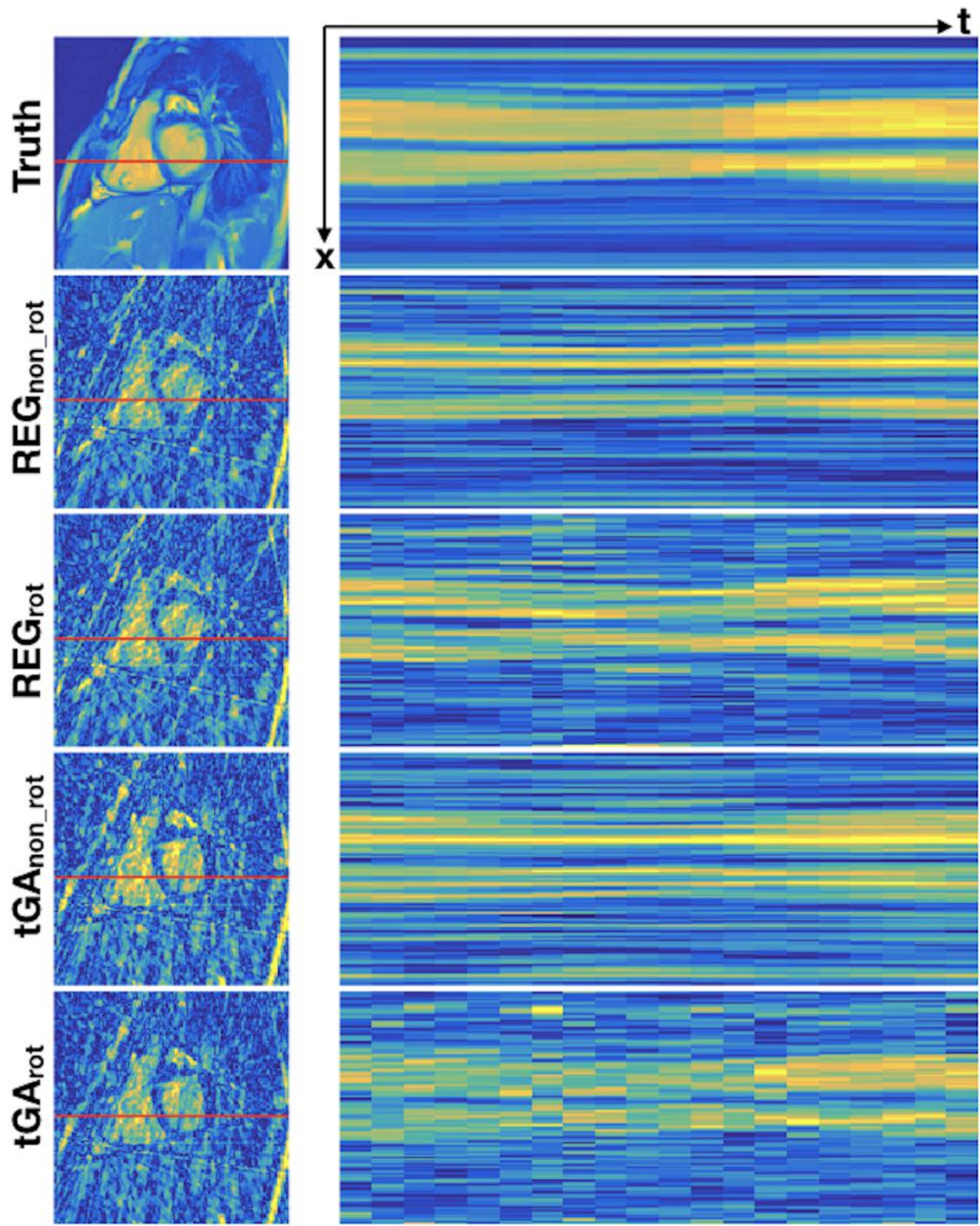

## Residual U-Net Architecture and Training

Four residual U-Nets were trained; one for each of the radial sampling patterns. The synthetic training data for each U-Net consisted of paired artifact-free 'ground truth' magnitude images and the corresponding undersampled, artifact-contaminated images (different for each sampling pattern). The architecture of the residual U-Net relied on a multi-scale decomposition of the input image and a skip connection at each scale, see Figure 3 for the chosen architecture. Each convolutional layer had a filter size of 3x3x3 and was equipped with a rectified linear unit as nonlinearity, except the last layer that produced the residual update. The filters were equally weighted in spatial and temporal domain and hence no directions were favoured in the training process. The output of the network was the input artifact-contaminated image with the residual update and the result was projected to positive numbers by a rectified linear unit to enforce non-negativity.

**Figure 3** *The chosen residual U-Net structure for spatio-temporal deep de-aliasing. The input is given by the aliased reconstruction from undersampled data. The numbers on top of the blue bars denote the number of channels for each layer. The resolution for each multilevel decomposition is shown in gray on the left. Each convolutional layer is equipped with a Rectified Linear Unit as nonlinearity, given by ReLU(x)=max(x,0).*

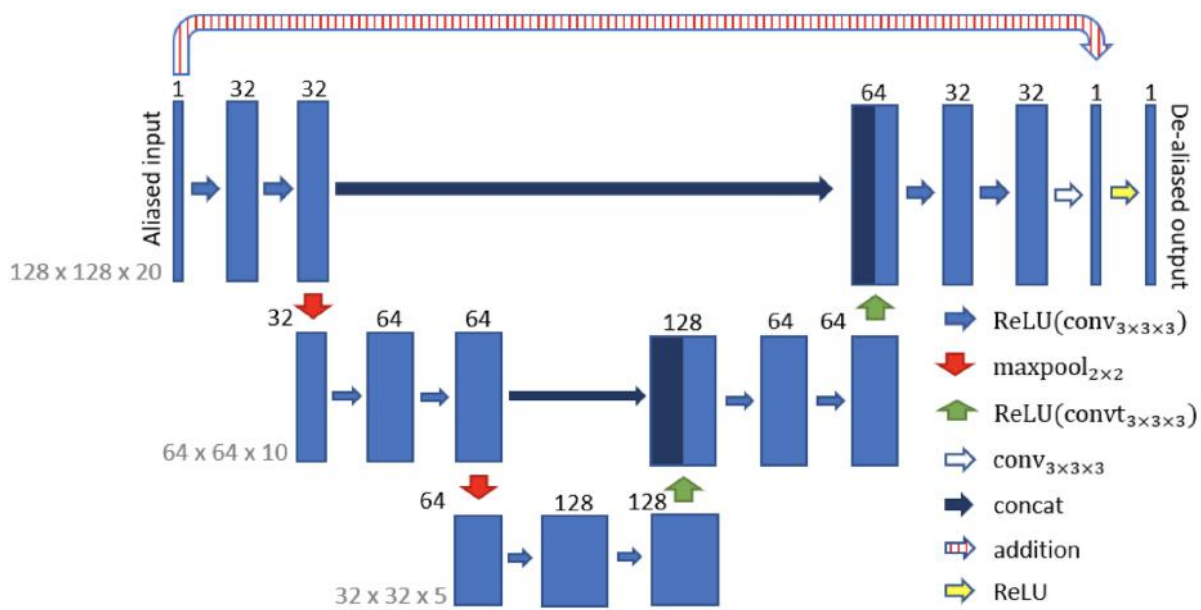

Implementation and training of the residual U-Net was done in Python with TensorFlow (14). We minimised the $\ell^2$-loss of the reconstructed volume to the desired ground truth. The training was done for 350 epochs with the Adaptive Moment Estimation algorithm (ADAM) (15), with an initial step size of $10^{-3}$ and batches of 8 samples. The total training time for each network took 12 hours on a Titan XP GPU with 12Gb memory.

**Evaluation of Synthetic Test Data**

The performance of the four separately trained networks (one for each sampling pattern) were tested using a synthetic test data set. This data set was created from BH-bSSFP data, previously acquired in 25 children and adults with paediatric heart disease or CHD (mean age: 24.3±13.3 years, range: 8 to 64 years, male: 13) that were not used to create the synthetic training data – full list of diagnoses in Supporting Information S2. Preparation of the synthetic test data was the same as for the training data. The artifact-contaminated magnitude data for each of the sampling patterns were input into their correspondingly trained networks. The output data were compared quantitatively and qualitatively to the ground truth images. Quantitative assessment was carried out using root mean square error (RMSE) and Structural Similarity Index (SSIM). The results of these analyses were averaged over the cardiac phases and slices for each patient. Qualitative assessment was performed using all cardiac phases from the middle slice of each patient. The output images for each sampling pattern were compared to the corresponding ground truth images by a CMR specialist (V.M., with 15 years' experience) and evaluated for: i) Absence or deformation of papillary muscles or trabeculations, ii) Poor motion fidelity, and iii) Blurring of the endocardial border.

**Assessment of Robustness**

The chosen protocol parameters and the level of noise in the artifact-contaminated images may affect the performance of the final trained network. The robustness of the resultant $tGA_{rot}$ network was assessed by modifying the synthetic real-time data, in terms of: i) signal-to-noise ratio (SNR), ii) acceleration factor, iii) image cropping.

The effect of SNR was assessed by artificially adding Gaussian noise to the artifact-contaminated tGA$_{rot}$ synthetic test images, varying the resultant SNR between 20 dB to 10 dB. To assess the effect of acceleration factor, the synthetic real-time data was undersampled with acceleration from 10x to 16x. The robustness to position of image cropping, was assessed by shifting the cropped region by 0/±4/±8/±12 pixels, in the x and/or y direction (giving 48 shifted regions, for each subject). Resultant images were compared using RMSE and SSIM.

**Actual In-vivo Study**

Actual real-time data was prospectively acquired in 10 children and adults with paediatric or CHD referred to our centre for clinical CMR (mean age: 33.6±16.8 years, range: 16 to 64 years, male: 3) – full list of diagnoses in Supporting Information S2. Exclusion criteria were: i) Inability to breath hold, ii) a single ventricle, and ii) Arrhythmia. All patients were imaged on a 1.5T MR scanner (Avanto, Siemens Medical Solutions, Erlangen, Germany) with VCG gating. Real-time undersampled radial data was acquired, as described below. In addition, reference standard assessment of ventricular volumes was performed using standard BH-bSSFP cine imaging in the short axis (acquisition parameters same as parameters of the training data; matrix ~240x204, FOV ~320x265 mm, phase partial-Fourier 6/8, spatial resolution ~1.3x1.3 mm, slice thickness ~10 mm, Bandwidth ~940 Hz/Pixel, Flip angle ~70$^o$, TE/TR ~1.2/2.4 ms temporal resolution ~32 ms, reconstructed cardiac phases 40). Ten to 15 contiguous slices were acquired in the short axis to ensure coverage of the whole ventricular volume, with one slice acquired per breath hold (~6 seconds).

In addition, actual real-time data was prospectively acquired in one healthy volunteer to assess the effect of rapid or large respiratory motion on the performance of the final trained network. All imaging parameters were the same as above. Real-time data were acquired during breath-holding, normal breathing and rapid, deep breathing.

*Real-time Sequence:* The new real-time sequence was based on an in-house single-shot multi-slice 2D radial bSSFP technique. Imaging parameters were: matrix 192x192, FOV 320x320 mm, spatial resolution 1.67x1.67 mm, slice thickness 10 mm, Bandwidth 1240 Hz/Pixel, Flip angle 67°, TE/TR 1.4/2.8 ms. Consecutive spokes were separated by the tiny golden angle and 14 spokes were used to reconstruct each frame (no temporal overlap between frames). This represented an undersampling factor of 13x, based on 182 spokes being required to fully sample k-space (16). The resultant temporal resolution was 36.4 ms. Acquisition of each slice required two R-R intervals; the first R-R interval being used to reach the steady state and the second for data acquisition. Ten to 15 contiguous slices were acquired in the short axis and after acquisition the imaging plane was moved to the next contiguous slice. Image acquisition was performed during free breathing, using two spine coils and one body-matrix coil (giving a total of 12 coil elements).

*Reconstruction of Real-time Data:* Creation of artifact-contaminated images from the undersampled radial data was performed online in Siemens reconstruction environment (ICE VB17, two AMD Opteron processors, clockspeed: 3.2GHz, cache: 6MB). The k-space data was gridded, Fourier transformed into image space, followed by weighted sum over the coil elements. The resultant artifact-contaminated images were processed in MATLAB similarly to the test data; cropped to 128x128 matrix, followed by temporal interpolation to 20 frames. The resultant images subsequently had the artifact suppressed using the previously trained $tGA_{rot}$ residual U-Net.

The undersampled radial data was also reconstructed using the CS algorithm, GRASP (Golden angle Radial Acquisition with temporal total variation SParsity regularization) (17). In order to provide an unbiased comparison in terms of image quality and reconstruction time, we used the Berkeley Advanced Reconstruction Toolbox (BART) (18) as it supports parallel computation using GPUs. This toolbox uses ADMM for temporal total variation constraints (19). The number of iterations was set to 50 as a good trade-off between reconstruction speed and reconstruction accuracy (determined from preliminary experiments described in Supporting Information Figure S4) Coil sensitivity information was estimated from the k-space centre (20). The regularisation level was selected empirically (to 0.025) based on the first 5 patients to minimise artifacts without producing too much temporal blurring. The GRASP algorithm was performed on the same computer as the residual U-Net (Titan XP GPU with 12Gb memory).

**Analysis of In-vivo Data**

Image quality and ventricular volumes were compared between the gold-standard BH-bSSFP images, and the real-time radial data reconstructed with the GRASP algorithm and the deep artifact suppression residual U-Net.

*Image Quality:* Qualitative image scoring was performed by a CMR specialist (V.M.). For all patients, the mid-ventricular short-axis cine images from each technique were viewed in a random order. Each cine loop was graded on a 5-point Likert scale in four categories: sharpness of endocardial border (1=non-diagnostic, 2=poor, 3=adequate, 4=good, 5=excellent), temporal fidelity (or blurring) of wall motion (1=non-diagnostic, 2=poor, 3=adequate, 4=good, 5=excellent) and residual artifacts (1=non-diagnostic, 2=severe, 3=moderate, 4=mild, 5=minimal).

Quantitative edge sharpness was calculated by measuring the maximum gradient of the normalized pixel intensities across the border of the septum, as previously described (21). To reduce noise, which results in artificially high gradients (representing sharp edges), the pixel intensities were fit to a tenth order polynomial, before differentiation. Edge sharpness was calculated in six positions across the septum, for all cardiac phases, and the average value was used for comparison.

*Ventricular Function:* Quantification of left and right ventricular volumes was calculated for each technique. The end-diastolic and end-systolic phases for each ventricle were chosen through visual inspection of the mid-ventricular cine. At end diastole and systole, all ventricular sections were manually segmented by a CMR specialist (V.M.). The endocardial border was traced using the OsiriX open source DICOM viewing platform (Osirix v.9.0, OsiriX foundation, Switzerland) (22). The papillary muscles and trabeculae were included in the blood-pool. End-diastolic volume (EDV), end-systolic volume (ESV) and ejection fraction (EF) were calculated as previously described (23).

## Statistics

Statistical analyses were performed by using STATA software (STATA SE, v.14.2). Qualitative image scores for synthetic test data are expressed as percentage of cases where the artifact was visible. Comparisons of continuous variables was performed using one-way repeated measures analysis of variance (ANOVA) with post hoc testing using Bonferroni correction for significant results. Comparison of ratio data (image scoring of synthetic test data) was performed with the chi-squared test and comparison of ordinal data (image scoring of clinical data) was performed using the Wilcoxon signed-rank test. For assessment of agreement of ventricular volumes and function, the BH-bSSFP data was used as the reference standard for Bland-Altman analysis. A p-value of less than 0.05 indicated a significant difference.

# RESULTS

## Synthetic Test Data

Figure 4 shows image quality of the synthetic test data for the different sampling patterns (corresponding movie can be seen in Supporting Information Video S5). Table 1 shows the RMSE and SSIM results for the different radial sampling patterns. The $tGA_{rot}$ had significantly lower RMSE (better reconstruction accuracy) than all of the other sampling patterns (p<0.0001) and significantly higher SSIM (better reconstruction accuracy) than all the other sampling patterns (p<0.0001). Qualitative measures of the images from the different sampling patterns are also shown in Table 1. The $tGA_{rot}$ had significantly lower proportions of absent/deformed papillarys/trabeculations, poor motion fidelity and blurred endocardial borders compared to $tGA_{no\_rot}$ and $REG_{no\_rot}$ (p<0.0001). In addition, $tGA_{rot}$ had significantly lower proportions of absent/deformed papillarys/trabeculations (p=0.018) and poor motion fidelity (p=0.045) than $REG_{no\_rot}$.

**Figure 4** *Examples images from two 'synthetic real-time' test data sets reconstructed from the four different sampling patterns, using correspondingly trained residual U-Net's. Pt1 shows total loss of some papillary muscles (indicated with arrow) compared to the truth data, with all trajectories except tGA$_{rot}$. Pt2 shows blurring across the ventricle, particularly seen with the non-rotating trajectories. (see* Supporting Information Video S5 *for corresponding movie)*

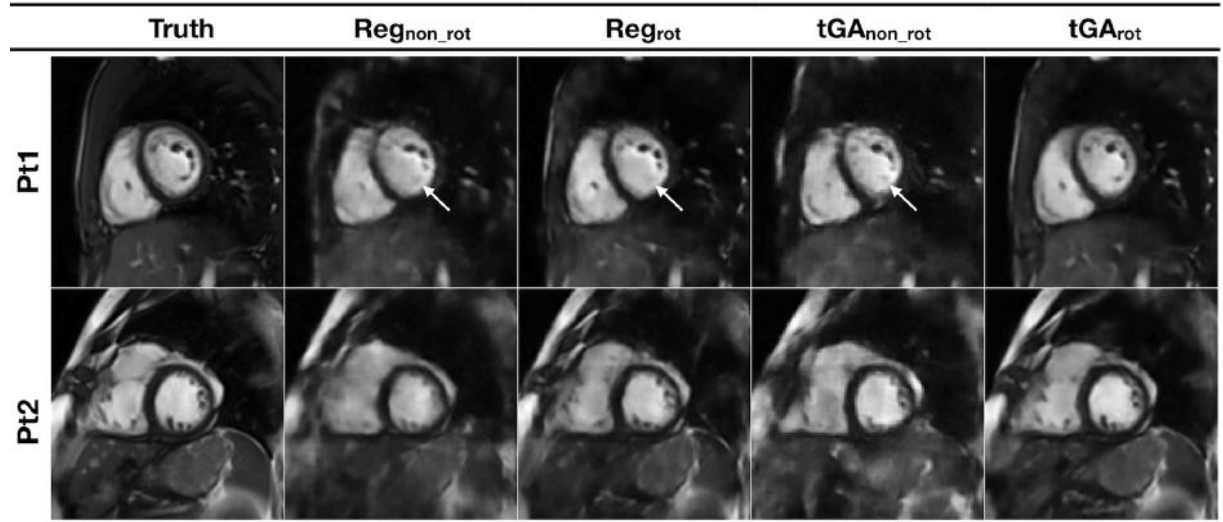

**Table 1:** *Assessment of different radial sampling patterns. Quantitate measures: RMSE and SSIM calculated with the 'truth' images. Qualitative measures: percentage of case with visible artifacts: Pap: Absence or deformation of papillary muscles or trabeculations, Motion: Poor motion fidelity, and Blur: Blurring of the endocardial border.*

|  | REG$_{no\_rot}$ | REG$_{rot}$ | tGA$_{no\_rot}$ | tGA$_{rot}$ |
|---|---|---|---|---|
| **RMSE** (x10$^{-2}$) | 8.0 ± 1.5 * | 4.9 ± 1.0 * | 8.4 ± 1.6 * | 4.2 ± 1.3 |
| **SSIM** | 0.64 ± 0.04 * | 0.83 ± 0.03 * | 0.63 ± 0.05 * | 0.87 ± 0.03 |
| **Pap** | 92% * | 20% * | 88% * | 0% |
| **Motion** | 48% * | 24% * | 48% * | 4% |
| **Blur** | 100% * | 68% | 100% * | 52% |

*\* Values are significantly different from tGA$_{rot}$ (p<0.05)*

*Assessment of robustness:* For the tGA$_{rot}$ network, as SNR decreased from 20 dB to 10 dB, the SSIM decreased by 6.8% (0.862 at 20 dB, 0.794 at 10 dB, and 0.868 with no additional noise) and the RMSE increased by 1.3% (0.043 at 20 dB, 0.056 at 10 dB, and 0.042 with no additional noise) - see Supporting Information Figure S6. As acceleration factor increased from 10x to 16x, SSIM decreased by to 6% (0.89 at 10x, and 0.83 at 16x) and RMSE increased by 0.9% (0.041 at 10x, and 0.050 at 16x) - see Supporting Information Figure S7. Unsurprisingly, the 13x undersampled data performed better than this trend, as the network was trained on 13x undersampled data. Altering the position of image cropping had little effect on SSIM and RMSE. The lowest SSIM was 0.858 at a shift of +12 pixels in x and y, and was 1.04% less than with no shift applied. The highest RMSE was 4.53 x$10^{-2}$ also at a shift of +12 pixels in x and y, and was 0.34% greater than with no shift applied. See Supporting Information Figure S8 for all results.

**Prospective In-vivo Study**

Real-time data was successfully acquired and reconstructed (GRASP and residual U-Net) in all 10 patients. On average, there were ~12 slices per patient, resulting in 122 3D data sets. Reference standard BH-bSSFP data was also successfully acquired in all patients. Total acquisition time for BH-bSSFP stack was 279±65 s, and for the real-time radial bSSFP was 18±3 s.

Reconstruction Time: Overall reconstruction times for all slices (~12 slices, from raw data file until completion of reconstruction) was ~5.6x faster for deep artifact suppression compared to compressed sensing (~22.0 s vs ~111.4 s).

The compressed sensing reconstruction took ~3.4 s to read the raw k-space data file (containing all phases, for all slices, with 12 coil elements), and the GRASP algorithm (including calculation of trajectories and coil sensitivity information) took ~8.9 s per slice.

For the deep artefact suppression, creation of artifact-contaminated DICOM images took ~14.8 s for all slices (~1.3 s per slice, including reading of the raw k-space data file, gridding of data, coil combination and saving to DICOM). Deep artifact suppression of these images took ~7.2 s for all slices, including ~5.4 s to read all DICOM files (all phases, for all slices), before initialisation of variables for each slice and running the residual U-Net (~0.15 s per slice).

*Image Quality:* Representative images are shown in Figure 5 and the corresponding movie can be seen in Supporting Information Video S9. The residual U-Net reconstruction of the real-time data had better image quality than the same data reconstructed using GRASP. In particular, there is still significant residual aliasing seen in the GRASP reconstruction, as well as greater temporal blurring (as seen in the x-t plots of Figure 5). There are some subtle differences between the real-time reconstructions and the BH-bSSFP images (i.e. reduced pericardial fat). These are probably due to minor slice position variations between the free breathing and breath-holding acquisitions. These differences between the U-Net and GRASP reconstructions are reflected in the qualitative image scoring (Table 2), with the residual U-Net reconstruction having similar myocardial delineation, motion fidelity and artifact scores to the BH-bSSFP images, and superior scores to the GRASP reconstructed data ($p<0.05$). Quantitative edge sharpness (Table 2) was higher in the BH-bSSFP images, followed by the residual U-Net reconstruction and then GRASP reconstruction, with all comparisons being significant ($p<0.007$).

**Figure 5** *Example image quality in two prospective patients, shown in peak systole and peak diastole, as well as an x-t plot, from the BH-bSSFP sequence and the real-time radial sequence reconstructed with GRASP and the residual U-Net. (see* Supporting Information Video S9 *for corresponding movie)*

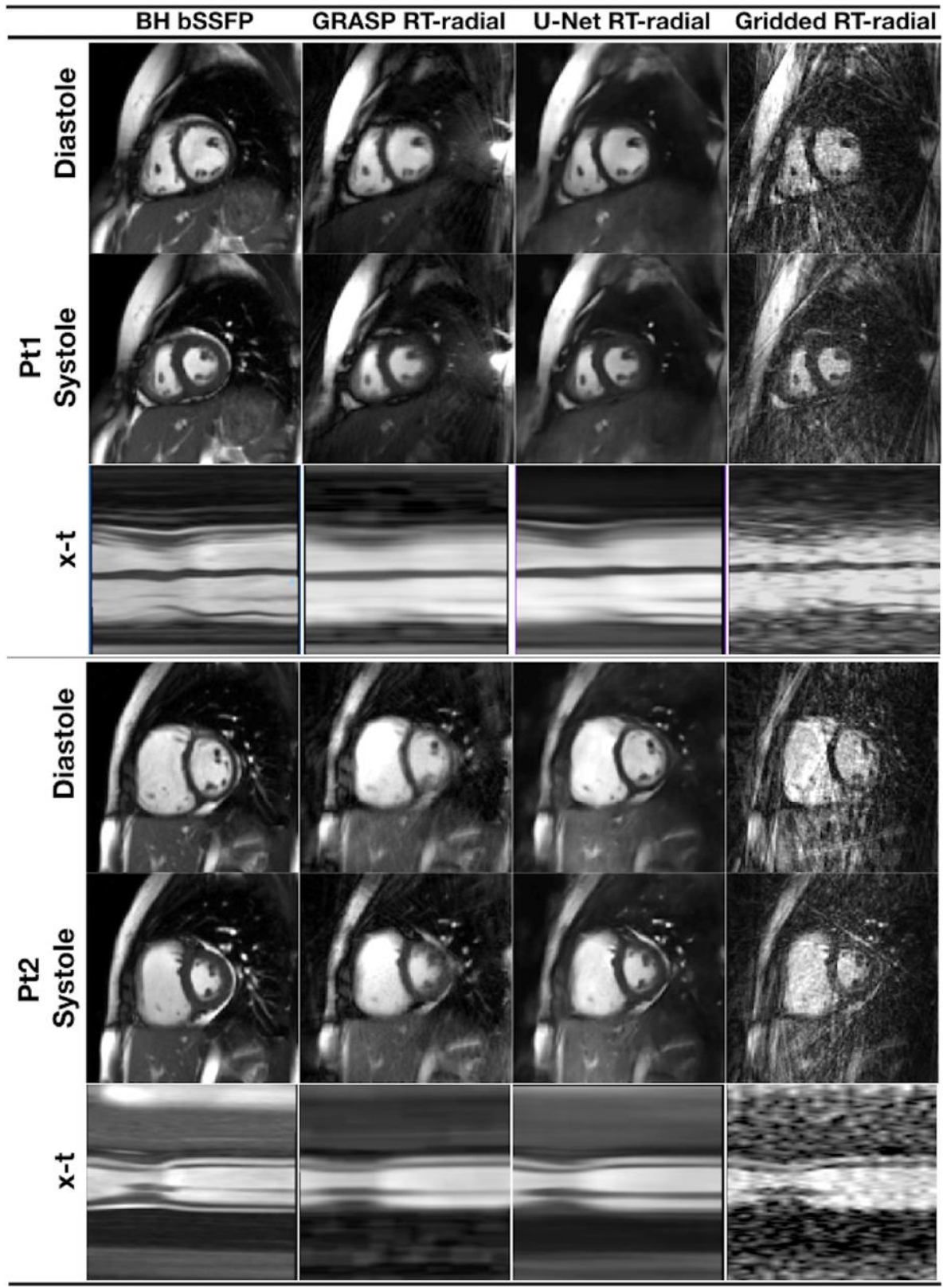

**Table 2** *Quantitative and qualitative image quality results from prospective in-vivo data, comparing the BH-bSSFP sequence, to the real-time radial sequence with GRASP and residual U-Net reconstructions. Quantitative image scoring: Edge Sharpness (ES) results. Qualitative image scoring (5-point Likert scale): Edge - sharpness of endocardial border, Motion - temporal fidelity (or blurring) of wall motion, Artifact - residual artifacts.*

|          | BH-bSSFP      | RT GRASP          | RT U-Net       |
|----------|---------------|-------------------|----------------|
| **ES**       | 0.66 ± 0.15   | 0.38 ± 0.08 *†    | 0.55 ± 0.14 *  |
| **Edge**     | 4.7 ± 0.45    | 3.4 ± 0.49 *†     | 4.2 ± 0.75     |
| **Motion**   | 4.8 ± 0.40    | 3.4 ± 0.49 *†     | 4.3 ± 0.46     |
| **Artifact** | 4.9 ± 0.30    | 3.8 ± 0.40 *†     | 4.6 ± 0.49     |

\* Values are significantly different from BH-bSSFP ($p<0.05$)

† Values are significantly different from RT U-Net ($p<0.05$)

Figure 6 shows images from the single volunteer under different respiratory conditions and the corresponding movie can be seen in Supporting Information Video S10. It can be seen that the U-Net reconstruction appears robust to increasing respiratory motion. This is in contrast to the GRASP reconstruction which is compromised during deep rapid breathing, particularly at end diastole.

**Figure 6** *Image quality in a single volunteer from the real-time radial sequence reconstructed with GRASP and the residual U-Net, under different respiratory conditions; breath-holding, free-breathing and deep, rapid breathing. (see Supporting Information Video S10 for corresponding movie).*

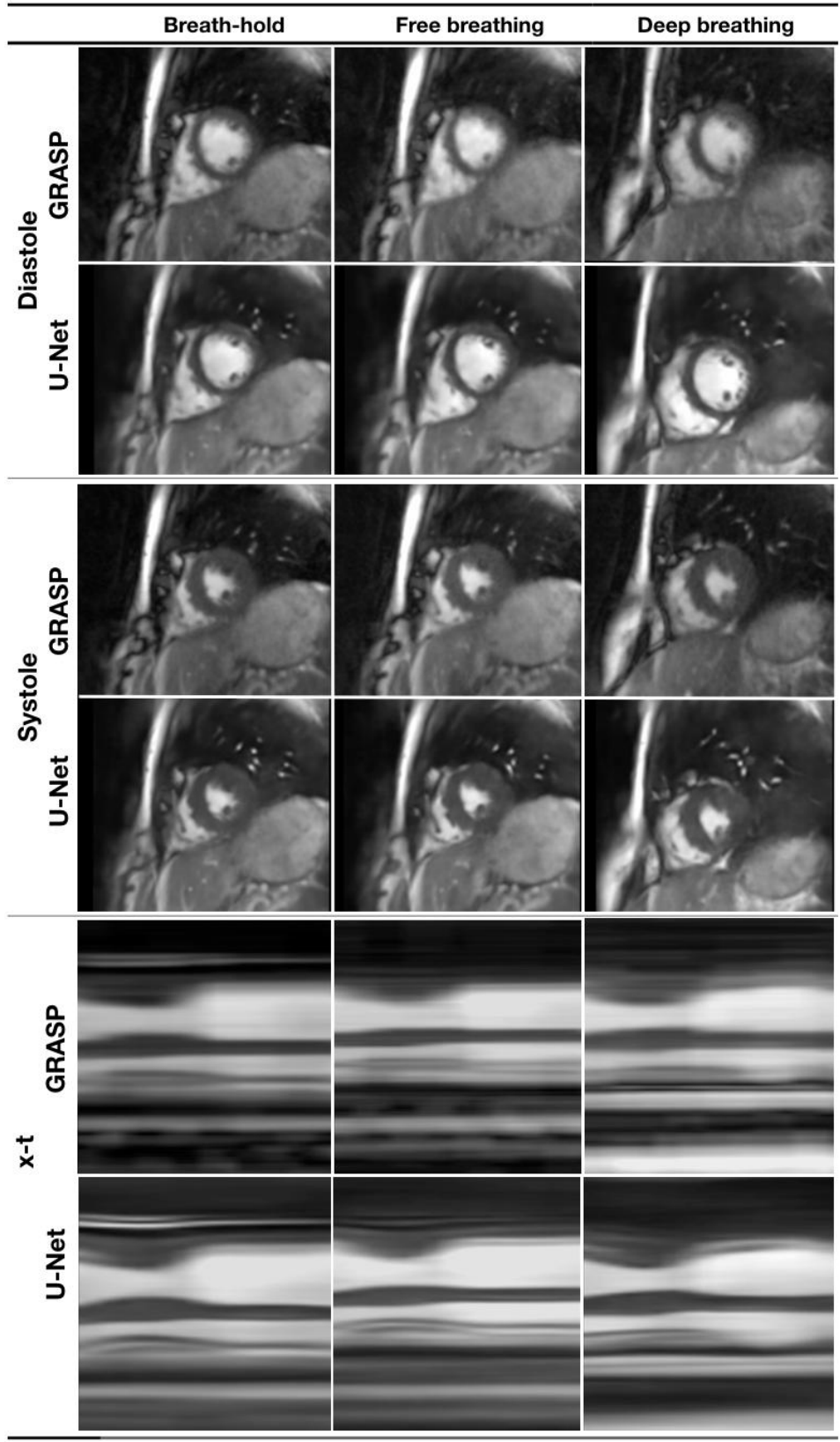

*Ventricular volume quantification:* Biventricular volumes and function are shown in Table 3 (with all Bland-Altman graphs for the LV in Supporting Information Figure S11 and for the RV in Supporting Information Figure S12). The limits of agreement for both the residual U-Net and GRASP reconstructions were similar. However, there were several significant biases with the GRASP reconstructed data including left ventricular ESV (p=0.04) and EF (p<0.001), and right ventricular EDV (p<0.001). The residual U-Net reconstructed data was only associated with a significant bias for right ventricular EDV (p=0.004), which was still less than the GRASP data (p=0.036). The Bland-Altman graphs for LV ESV, LV EF and RV EDV are shown in Figure 7.

**Table 3** *Ventricular volumes from prospective in-vivo data comparing the BH-bSSFP sequence, as well as the real-time radial sequence with GRASP and residual U-Net reconstructions. Left Ventricular (LV) and Right Ventricular (RV) volumes: EDV – end diastolic volume, ESV – end systolic volume, EF – ejection fraction.*

|  | Mean ± Standard deviation | | | Bias (Limits of agreement) | |
| --- | --- | --- | --- | --- | --- |
|  | **BH-bSSFP** | **RT GRASP** | **RT U-Net** | **RT GRASP** | **RT U-Net** |
| **LV EDV (mL)** | 148 ± 44 | 143 ± 44 | 151 ± 46 | -4.6 (-19.2 to 10.0) | 3.0 (-11.9 to 17.9) |
| **LV ESV (mL)** | 56 ± 27 | 60 ± 29* | 58 ± 29 | 4.1 (-8.0 to 16.2) | 1.9 (-7.9 to 11.7) |
| **LV EF (%)** | 64 ± 10 | 60 ± 11*† | 63 ± 11 | -4.1 (-9.5 to 1.3) | -0.3 (-4.1 to 3.5) |
| **RV EDV (mL)** | 213 ± 97 | 198 ± 89*† | 204 ± 92 | -14.9 (-33.4 to 3.6) | -8.6 (-23.2 to 6.0) |
| **RV ESV (mL)** | 92 ± 49 | 89 ± 48 | 91 ± 47* | -3.0 (-16.0 to 10.0) | -1.1 (-22.3 to 20.1) |
| **RV EF (%)** | 58 ± 7 | 57 ± 6 | 57 ± 6 | -1.5 (-9.1 to 6.1) | -1.4 (-9.5 to 6.7) |

*\* Values are significantly different from BH-bSSFP (p<0.05)*

*† Values are significantly different from RT U-Net (p<0.05)*

**Figure 7** *Bland-Altman plots showing ventricular volume results with significant biases. Comparison of ventricular volumes acquired with BH-bSSFP to real-time data reconstructed with GRASP and residual U-Net. Solid red line shows mean difference, dashed lines shows +/- 2 standard deviation.*

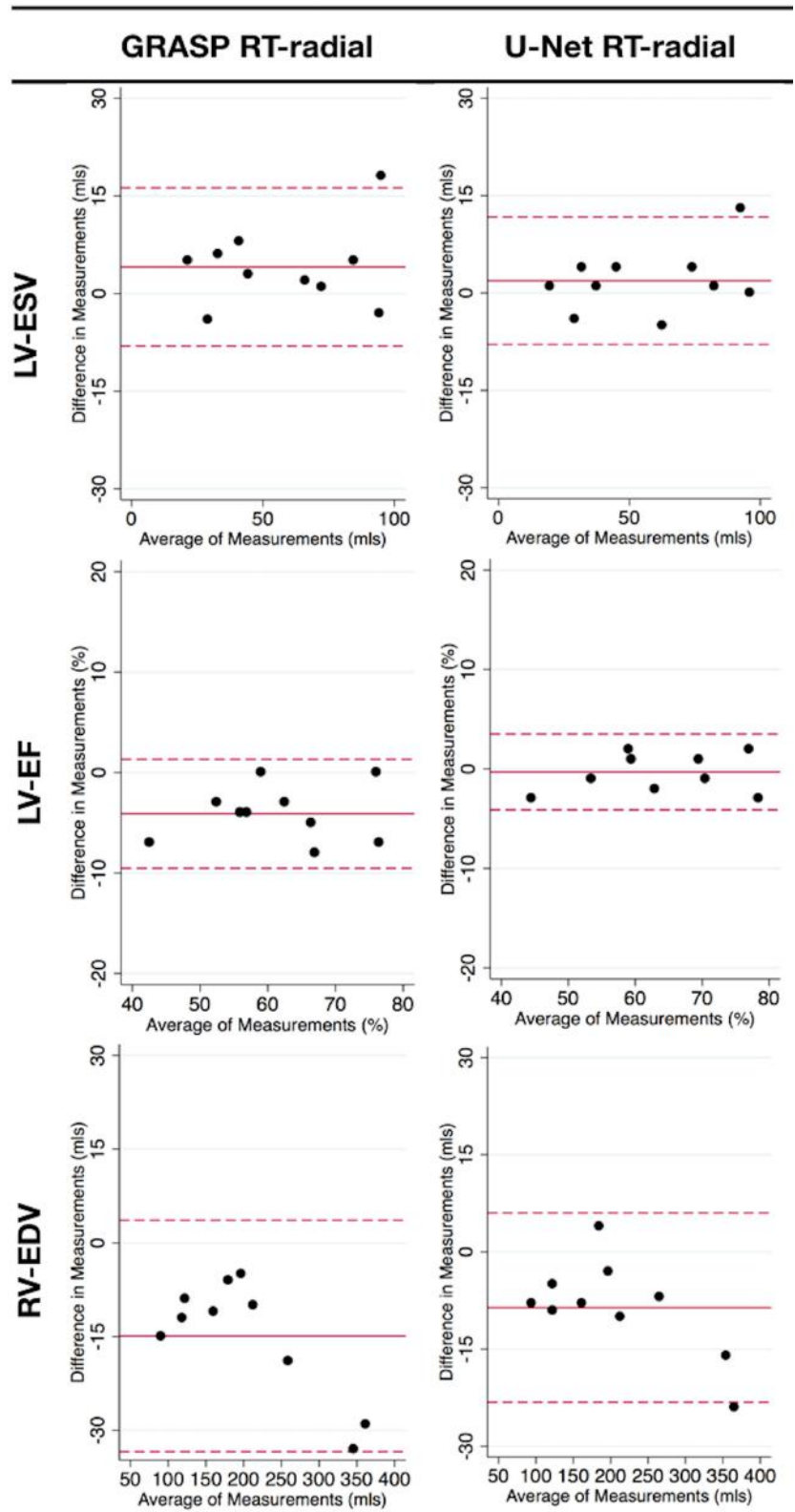

# DISCUSSION

The main findings of this study were: i) It is possible to train a 3D residual U-Net to remove artifacts (deep artifact suppression) in synthetic radially undersampled data created from previously acquired Cartesian BH-SSFP images, ii) The ability to perform deep artifact suppression on synthetic test data was determined by the simulated undersampling pattern, iii) Deep artifact suppression of actual radial real-time data acquired in new patients was successful with a residual U-Net trained using synthetic data, iv) Total reconstruction times were significantly shorter (>5x) using a residual U-Net compared to a conventional CS reconstruction, v) Image quality and measurement of ventricular volumes was superior using the residual U-Net compared to CS reconstructions of the same raw data.

## Deep Artifact Suppression and Synthetic Training Data

Machine learning, especially deep learning, relies on large amount of training data. Most clinical CMR services have large amounts of images stored from previously scanned patients. However, there are two problems with using this type of data to train deep learning algorithms capable of reconstructing real-time data. The first is that the CMR data is usually stored as magnitude images rather than complex raw data. Simple magnitude data is not suitable for training the iterative deep learning methods that have previously been used for CMR reconstruction (7,8,24-26). Therefore, in this study we chose to utilize a residual U-Net to remove aliasing from artefact contaminated gridded images. This approach has previously been shown to be successful in removing aliasing in retrospectively undersampled dynamic Cartesian CMR data (6). We have built on this previous work by using a much larger training data set and implementing non-Cartesian undersampling. The second problem with using previously acquired data is that it does not consist of paired ground truth and undersampled real-time data. Thus, retrospectively cardiac gated, Cartesian, breath-hold bSSFP images were used to create both synthetic training and test data. The advantage of using synthetic data is that the ground truth is known, the disadvantage is that errors during acquisition (such as gradient delays or trajectory errors) and respiratory motion (due to the free breathing acquisition for real-time imaging) cannot be corrected for. Nevertheless, we were able to show that this synthetic data could be used to train a residual U-Net to successfully perform deep artifact suppression.

Interestingly, the reconstruction accuracy of the synthetic test data was highly dependent on the radial undersampling pattern used. Non-rotating trajectories, where the same spokes were used in all frames, resulted in significant quantitative errors (as measured by RMSE and SSIM) and qualitative errors. Specifically, the images reconstructed for non-rotating trajectories were associated with missing or deformed papillary muscles/trabeculations in over 88% of synthetic test cases. Furthermore, the non-rotating trajectories also produced images with the worst motion fidelity and blurring of the endocardial border. In contrast, the continuously rotating strategies performed better, probably due to aliases having a less coherent structure through time. In particular, $tGA_{rot}$ produced aliases that appeared noise like and we believe this is the reason for the superior results seen with this sampling pattern. A full description of the association between the aliasing patterns and reconstruction accuracy are outside the scope of this paper. Nevertheless, the superior quality of $tGA_{rot}$ sampling for the synthetic test data confirmed that it was the right choice for the actual in-vivo part of this study.

## In-vivo Study

We demonstrated clinical translation of the residual U-Net by successfully performing deep artifact suppression on actual real-time images acquired in new patients. Importantly, because a $tGA_{rot}$ pattern was chosen it was possible to reconstruct the same raw data using a compressed sensing algorithm, namely the GRASP technique. One of the hypothetical advantages of a deep learning approach over CS is quicker reconstruction times. In this study, we found that residual U-Net reconstruction was >5x quicker than the GRASP reconstruction, when both were run on the same GPU. It should be noted that the reconstruction times for GRASP are almost linearly dependent on the number of iterations. Thus, reconstruction time could be easily shortened by simply reducing the number of iterations. However, this would reduce image quality of the GRASP reconstruction, which is already lower than the U-Net reconstruction. In addition, the U-Net evaluation time. Consequently, we believe that by accelerating reconstruction, deep learning approaches could play an important role in clinical translation of heavily undersampled real-time techniques.

In this study, we also found that the residual U-Net reconstructed images provided more accurate quantification of ventricular volumes compared to GRASP reconstructed images. The reason for this was probably better image quality, as reflected by better qualitative scores for endocardial border sharpness, temporal fidelity of wall motion, and residual artifacts. We believe that the superior image quality of the U-Net reconstruction was the result of using a high quality, large volume training data set. Conversely, the poorer CS reconstructions could be due to sub-optimal choice of the regularisation parameter. However, the presence of both residual artifacts and poor edge sharpness / temporal fidelity suggests that changing regularization would be of little benefit. In fact, the bias and limits of agreement of the GRASP reconstructed data were better than previous studies (4) using the same technique. This suggests that the GRASP implementation was adequately optimized. Thus, we believe that the residual U-Net reconstruction is superior and may allow more accurate measurement of ventricular volumes.

## Study Limitations

The main limitation is that the training data was acquired during a breath-hold, while the real-time data was acquired during free-breathing. This means that the network cannot 'learn' anything about respiratory motion, and may not be able to correct for this additional motion. However, we did we did perform an experiment in a single volunteer that demonstrated that the U-Net reconstruction was relatively robust to even large respiratory perturbations. This suggests that the network learns the structure of the noise to be removed, rather than the expected motion of the underlying structures. Nevertheless, further work is necessary to fully understand the effect of motion not apparent in the training data.

In this study, we also did not investigate the full diversity of CHD. Specifically, no patients with single ventricles were included in the training data set or the in-vivo study. Even though this first study shows that the trained network generalises well, the full generalisation capability of the network to varying motion and anatomies should be investigated in future work.

Another drawback of our implementation of a deep artifact suppression network was that all the training and subsequent data needed to be of the same size. However, when acquiring real-time data over one cardiac cycle, the number of frames is dependent on the patient's heart-rate. This meant that it was necessary to resample the aliased images through time, altering the temporal aliasing pattern of the acquired data. However, as both the training and actual real-time data were processed in the same way we do not believe that it is a significant limitation.

A further limitation of our approach was that the training and actual input data consisted of coil combined magnitude images, rather than raw multi-coil complex data. The main benefit of this approach was that previously acquired data that was easily retrievable from a conventional clinical image archive could be used for training. However, the absence of phase data in our approach may prevent optimum image restoration, in particular dealing with signal cancellation due to aliasing. Although this was not noticeable in the in-vivo data, this issue warrants further investigation in future studies. In addition, we did not investigate the effect of dark band artefacts in either the training data or actual input data on image quality. This study was performed at 1.5T and dark band artefacts were not a significant problem. Nevertheless, this issue must be investigated if our approach is to be considered for higher field strengths.

Finally, in this study we choose to minimise the $\ell^2$-loss and did not fully explore other loss functions such as $\ell^1$-loss. The $\ell^1$-loss function has been shown to provide superior training or improved results in image restoration (27), super resolution (28) and MRI image reconstruction (29). Therefore, we performed preliminary tests comparing the $\ell^1$-loss and $\ell^2$-loss using the tGA$_{rot}$ trajectory. Assessment of image quality from the trained networks using the synthetic test data, described above, demonstrated <0.01% difference in RMSE and <0.5% difference in SSIM (further information in Supporting Information Figure S13). This is similar to previous deep artifact removal studies, which have not shown large differences in reconstruction accuracy using different loss functions (6). Nevertheless, further investigation of different loss functions and other optimizations of learning would be desirable in the future.

## Conclusion

This paper demonstrates the potential of using a residual U-Net for deep artifact suppression of real-time radial data within a clinical setting. Once the networks have been trained, the reconstruction times are very short, making these techniques particularly appealing within busy clinical workflow. We have shown that ventricular volumes measured from images reconstructed using a residual U-Net are not statistically significantly different from the reference standard, cardiac gated, BH techniques. Thus, we believe that this technique may help with full adoption of real-time CMR in clinical practice. This would be particularly useful in children and sick patients who are unable to hold their breath.

# SUPPORTING INFORMATION

**Supporting Information Video S1**

*Movie showing residual flickering artifact when a 2D U-Net is applied to dynamic data. Left, shows images with residual aliasing, used as input to U-Net. Right, shows output from 2D U-Net. The flickering artifact is not seen with a 3D U-Net (2D plus time), as is used in this study.*

**Supporting Information S2**

*Diagnosis of patients used for training of the residual U-Net, creating of the synthetic test data, and for the prospective study.*

### Training data (N=250)

Repaired tetralogy of Fallot/pulmonary atresia with ventricular septal defect; 140

Normal; 38 (Family history cardiomyopathy)

Cardiomyopathy; 31

Coarctation of the aorta; 21

Aortic valve disease; 15

Repaired transposition of the great arteries; 4

Atrial septal defect; 1

### Synthetic Test data (N=25)

Repaired tetralogy of Fallot/pulmonary atresia with ventricular septal defect; 16

Aortic valve disease; 2

Coarctation of the aorta; 3

Cardiomyopathy; 2

Repaired transposition of the great arteries; 2

### Prospective real-time data (N=10)

Repaired tetralogy of Fallot/pulmonary atresia with ventricular septal defect; 4

Coarctation of the aorta; 2

Aortic valve disease; 2

Atrial septal defect; 2

**Supporting Information Figure S3**

*Flow diagram showing the steps taken to convert the retrospectively cardiac gated, BH-bSSFP data, to synthetic real-time data used to train the residual U-Net with.*

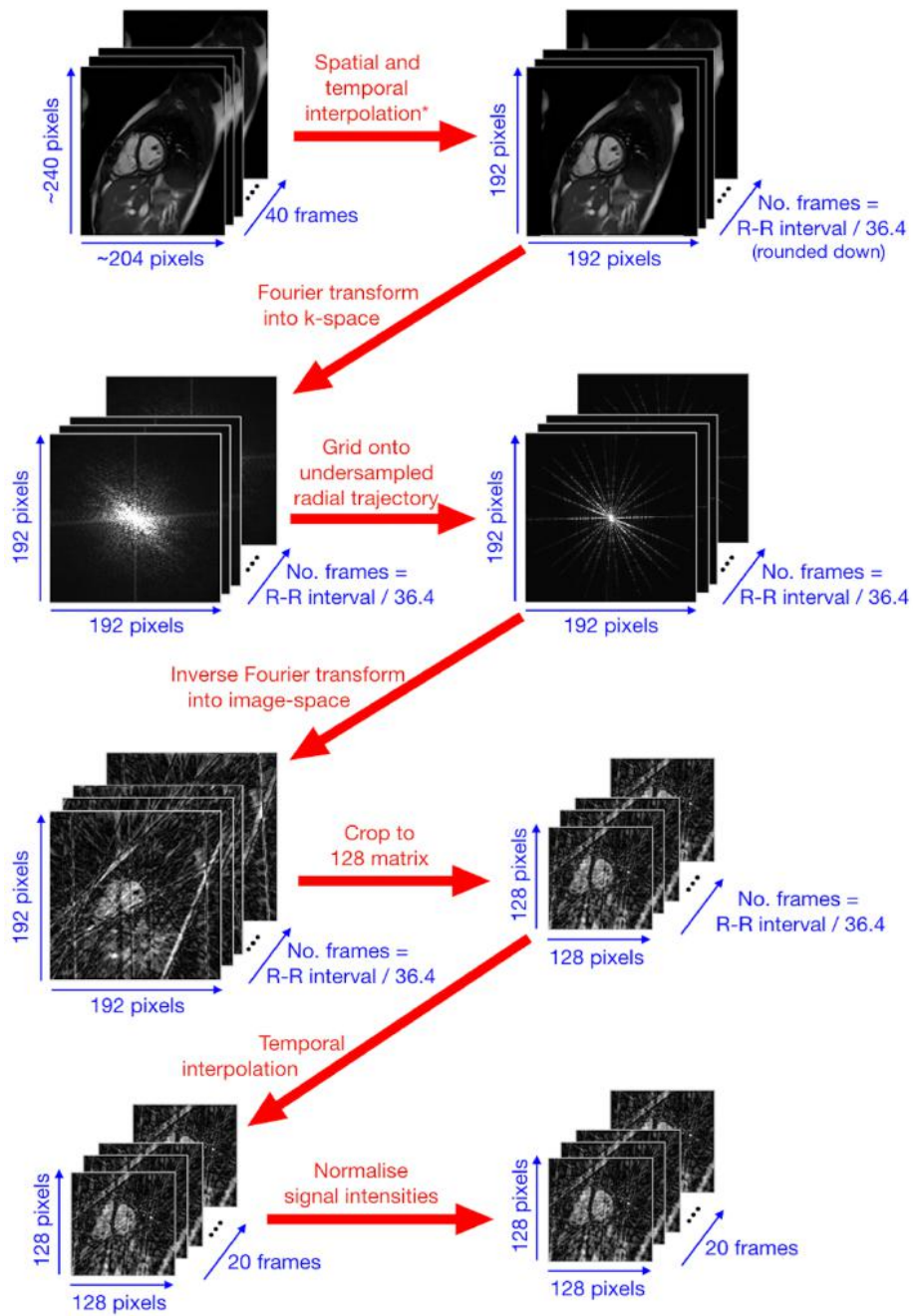

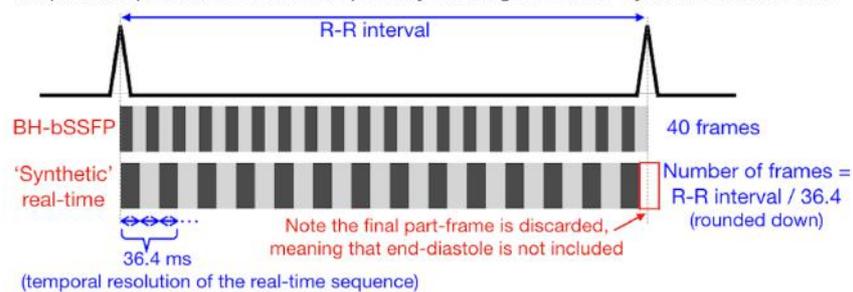

**Supporting Information Figure S4**

*Effect of number of iterations on GRASP reconstruction speed and GRASP reconstruction accuracy. Preliminary experiments showing; a) average reconstruction time per slice for different number of iterations (red line shows average time per slice for 50 iterations, as used in this study), b) Resultant image quality in one patient, for different numbers of iterations, c) Plot of difference in resultant images between different numbers of iterations (red line shows difference in images between 50 to 60 iterations), d) Difference images between numbers of iterations.*

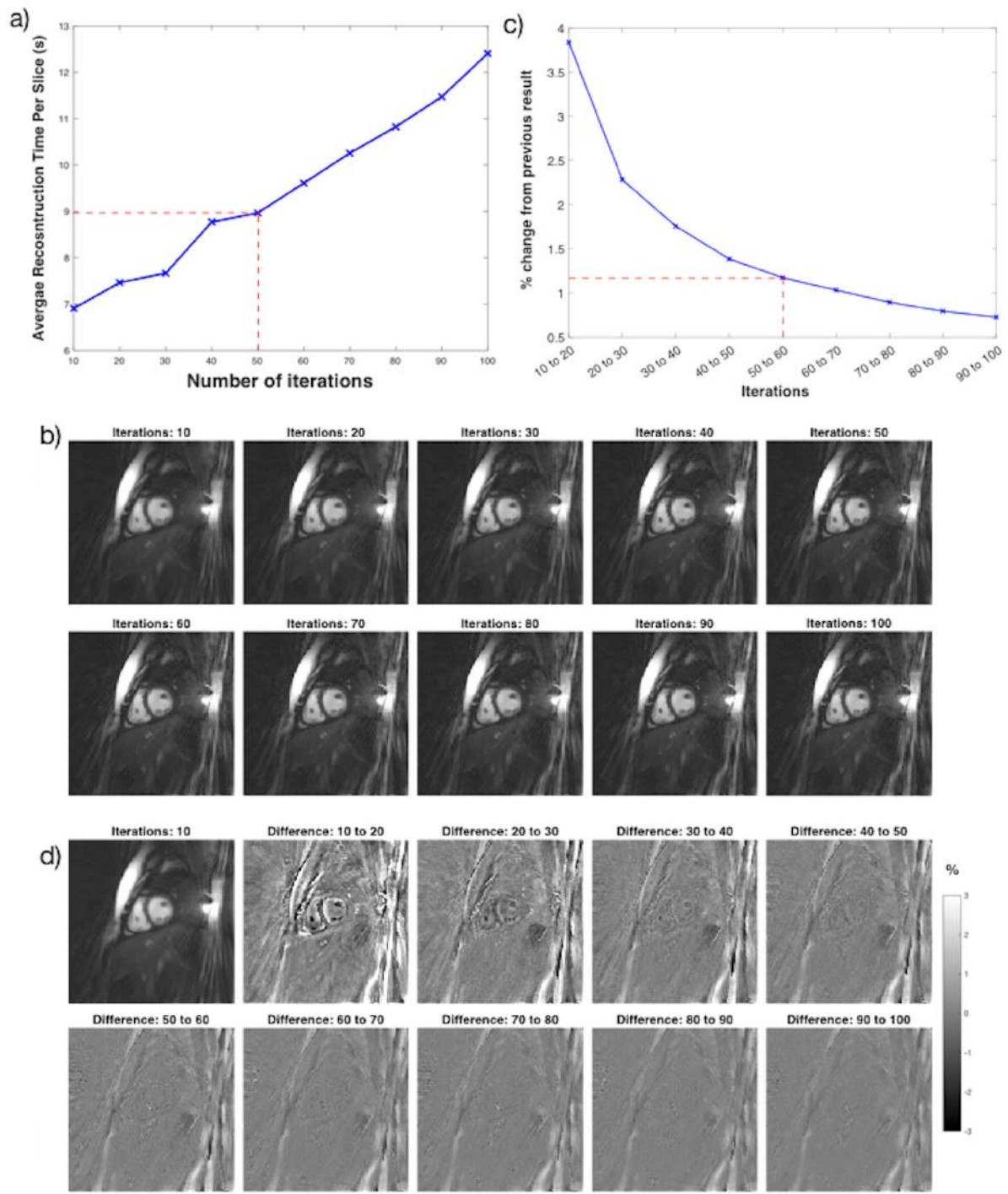

**Supporting Information Video S5**

*Example movies from two 'synthetic real-time' data sets reconstructed from the four different sampling patterns, using correspondingly trained residual U-Net's. Pt1 shows total loss of some papillary muscles compared to the truth data, with all trajectories except tGA$_{rot}$. Pt2 shows blurring across the ventricle, particularly seen with the non-rotating trajectories.*

**Supporting Information Figure S6**

*Assessment of Robustness: Effect of SNR. a) Average RMSE and SSIM results across all 222 test data sets, at different levels of SNR. b) Plot of RMSE at different SNR levels. c) Plot of SSIM at different SNR levels. b)c) Red line shows reference RMSE/SSIM with no additional noise. d) Example image quality from one patient, with input data from different SNR levels.*

### Effect of Signal-to-Noise ratio (SNR)

a) Average RMSE and SSIM results across all 222 test data sets, at different levels of artificial noise. b) Plot of RMSE at different SNR levels. c) Plot of SSIM at different SNR levels. b)c) Red line shows reference RMSE/SSIM with no additional noise. d) Example image quality from one patient, with input data from different SNR levels.

a)

| SNR (dB) | RMSE | SSIM |
|---|---|---|
| Original | 4.19 | 0.8684 |
| 20.0 | 4.31 | 0.8619 |
| 17.0 | 4.45 | 0.8544 |
| 15.2 | 4.60 | 0.8467 |
| 14.0 | 4.75 | 0.8388 |
| 13.0 | 4.91 | 0.8311 |
| 12.2 | 5.06 | 0.8236 |
| 11.5 | 5.21 | 0.8161 |
| 11.0 | 5.35 | 0.8087 |
| 10.5 | 5.48 | 0.8018 |
| 10.0 | 5.63 | 0.7942 |

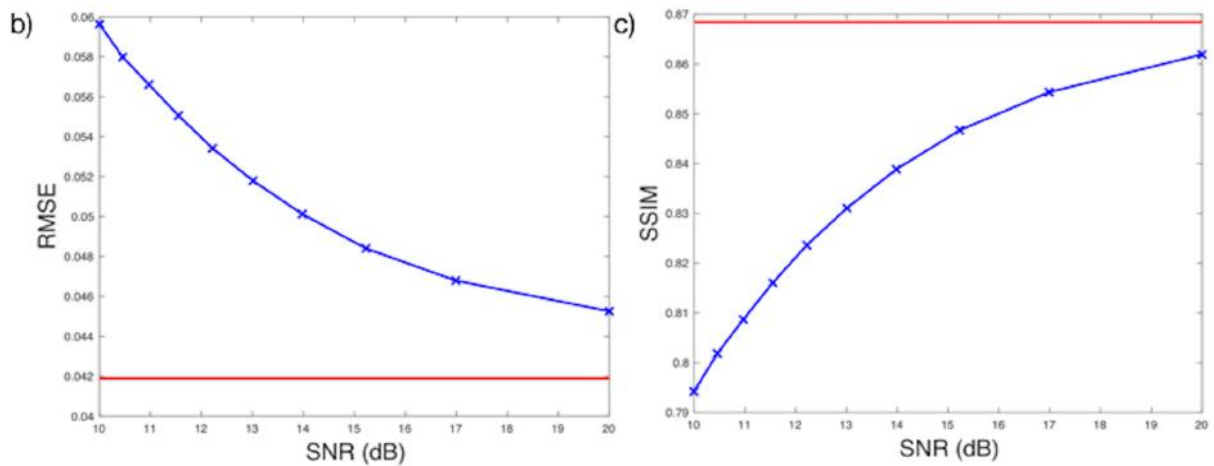

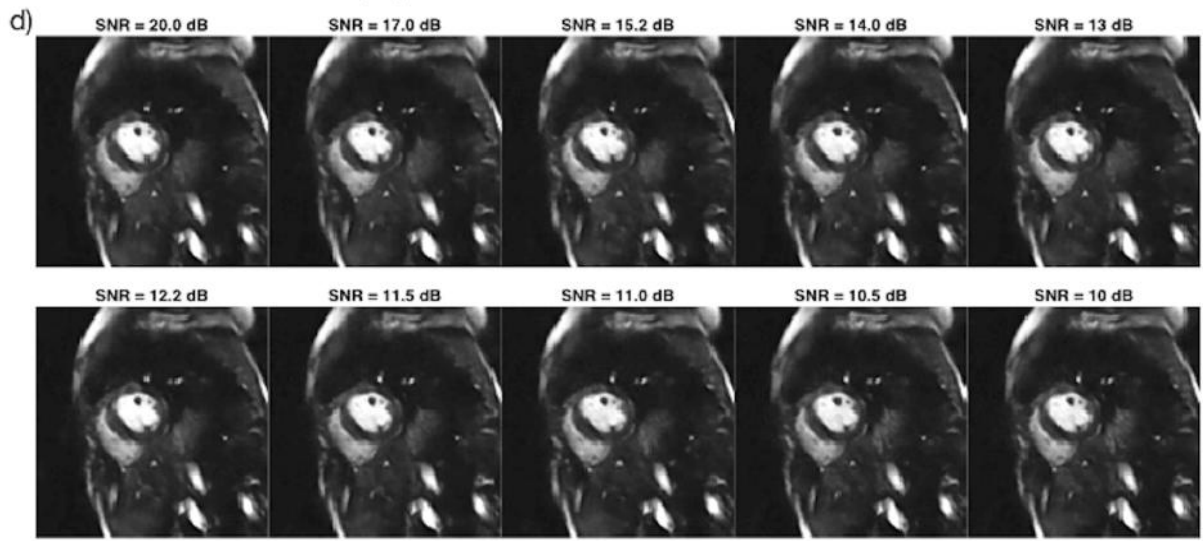

**Supporting Information Figure S7**

*Assessment of Robustness: Effect of Acceleration. a) Average RMSE and SSIM results across all 222 test data sets, at different acceleration factors. b) Plot of RMSE at different acceleration factors. c) Plot of SSIM at different acceleration factors. b)c) Red line shows reference RMSE/SSIM at 13x - this is the acceleration factor the network was trained with. d) Example image quality from one patient.*

**Effect of Acceleration**

a) Average RMSE and SSIM results across all 222 test data sets, at different acceleration factors. b) Plot of RMSE at different acceleration factors. c) Plot of SSIM at different acceleration factors. b)c) Red line shows reference RMSE/SSIM at 13x - this is the acceleration factor the network was trained with. d) Example image quality from one patient.

a)

| Acceleration | RMSE | SSIM |
| --- | --- | --- |
| 10x | 4.12 | 0.8915 |
| 11x | 4.36 | 0.8644 |
| 12x | 4.36 | 0.8644 |
| 13x | 4.19 | 0.8684 |
| 14x | 4.60 | 0.8472 |
| 15x | 4.60 | 0.8472 |
| 16x | 5.02 | 0.8276 |

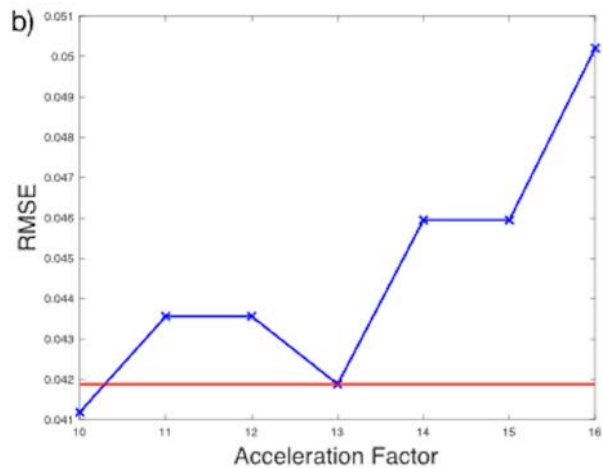
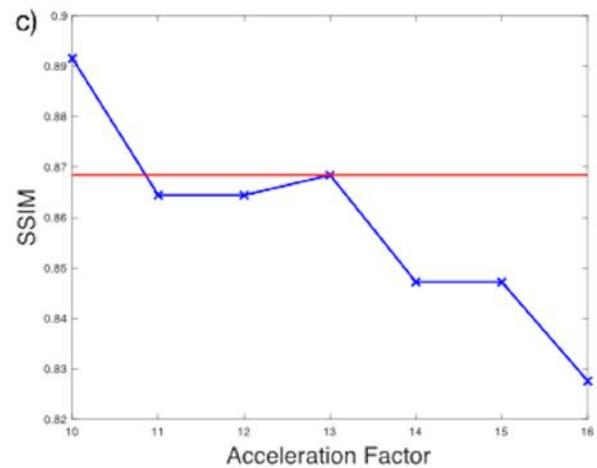
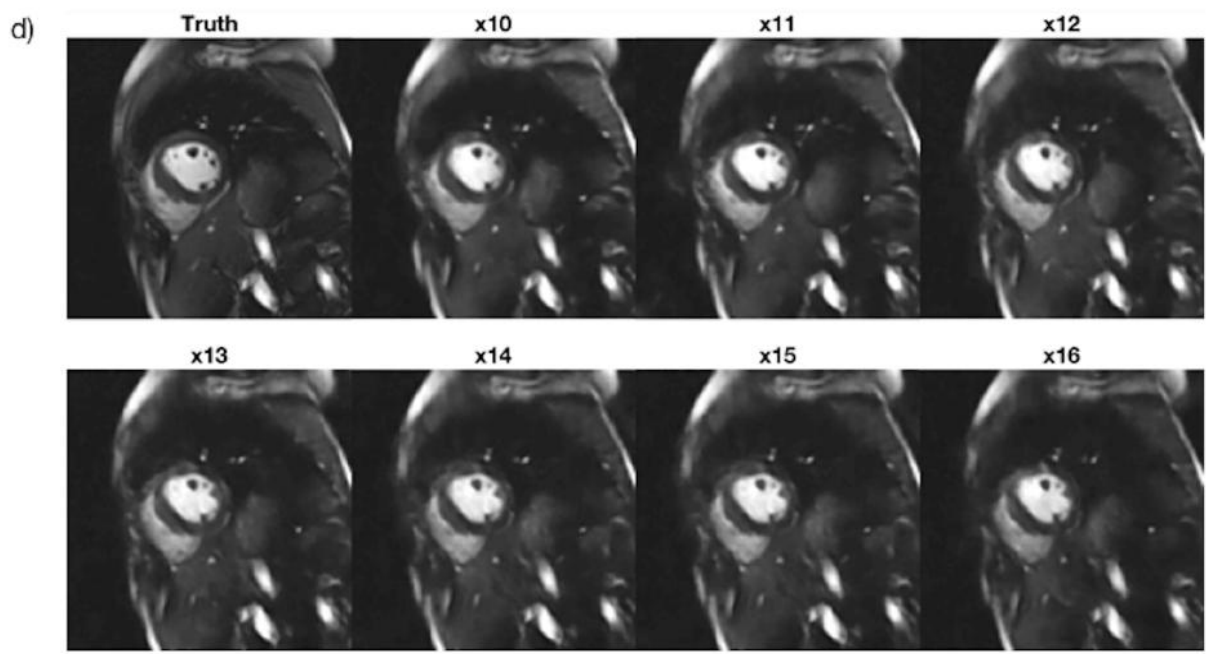

**Supporting Information Figure S8**

*Assessment of Robustness: Effect of shifting the cropping region. a) Average RMSE at different cropping positions in x and y, across all 222 test data sets. b) Plot of RMSE at different a cropping positions. c) Average SSIM at different cropping positions in x and y, across all 222 test data sets. d) Plot of SSIM at different cropping positions. e) Example image quality from one patient.*

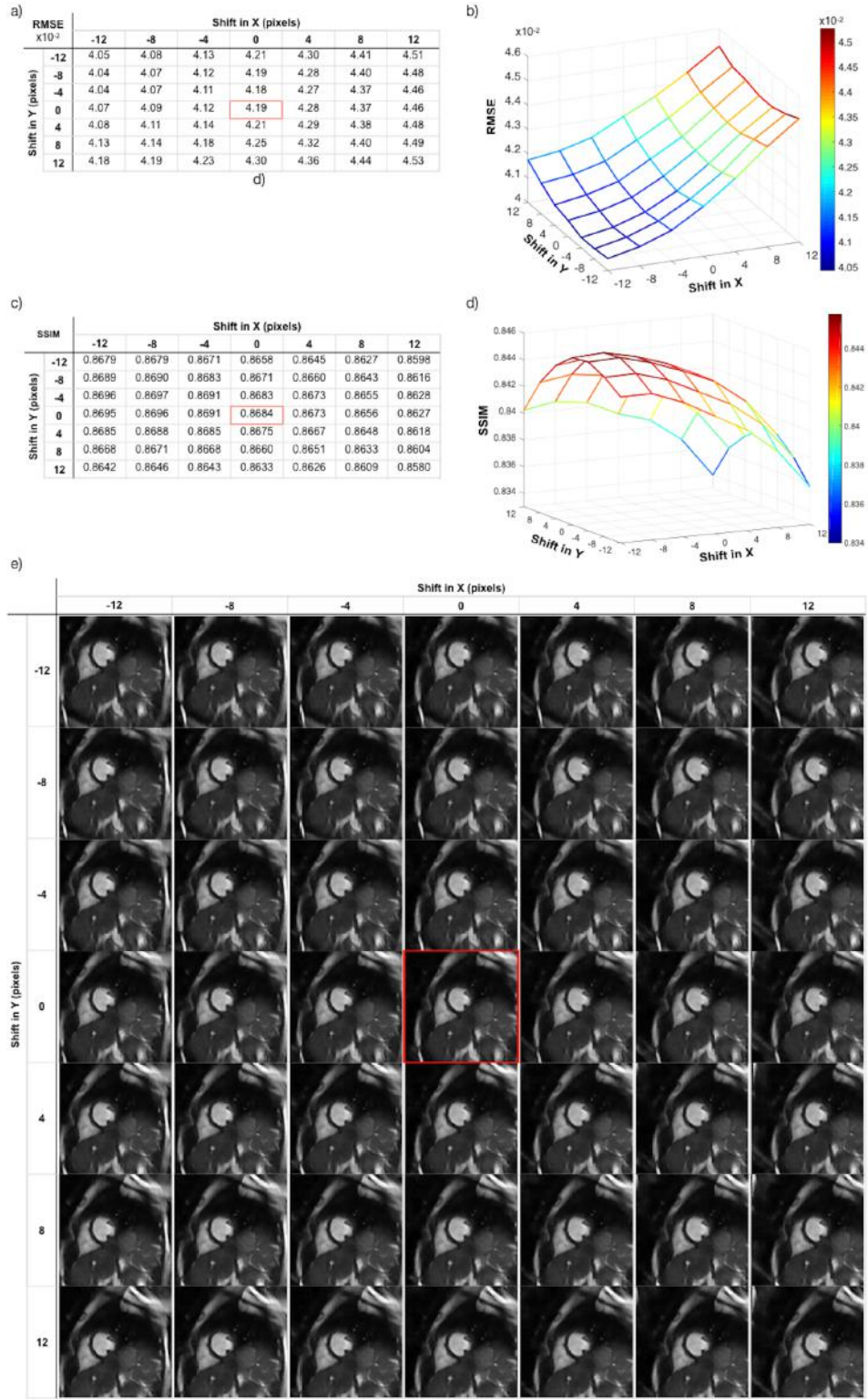

**Supporting Information Video S9**

*Example movies from two prospective patients, from the BH-bSSFP sequence and the real-time radial sequence reconstructed with GRASP and the residual U-Net.*

**Supporting Information Video S10**

*Example movies from a single volunteer under different respiratory conditions. Left to right; Breath-hold, Free-breathing, Deep breathing. Top row shows GRASP reconstruction, bottom row shows U-Net reconstruction.*

**Supporting Information Figure S11**

*Bland-Altman plot comparing left ventricular volumes acquired with BH-bSSFP to real-time data reconstructed with GRASP and residual U-Net. Solid red line shows mean difference, dashed lines shows +/- 2 standard deviation.*

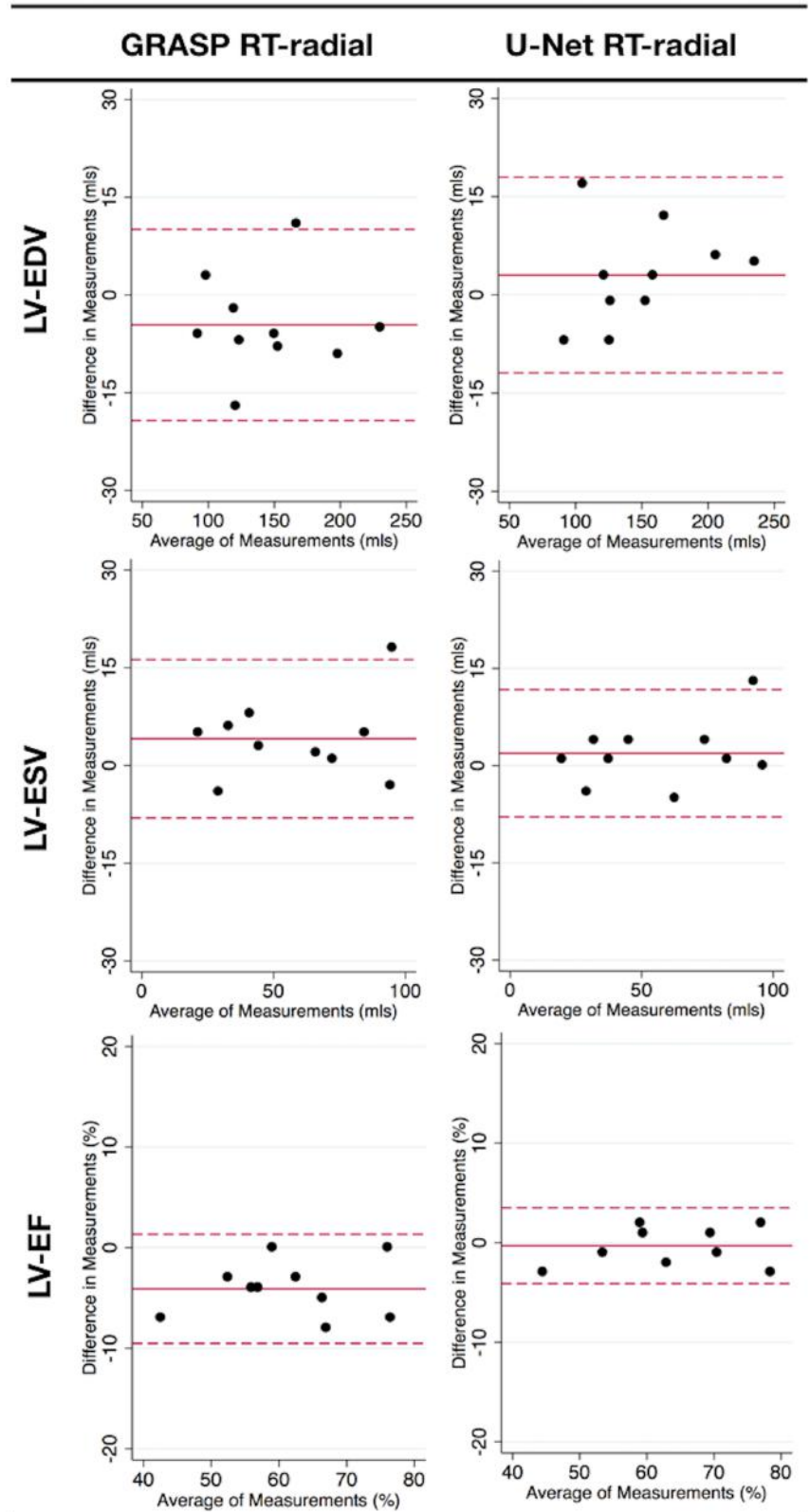

**Supporting Information Figure S12**

*Bland-Altman plot comparing right ventricular volumes acquired with BH-bSSFP to real-time data reconstructed with GRASP and residual U-Net. Solid red line shows mean difference, dashed lines shows +/- 2 standard deviation.*

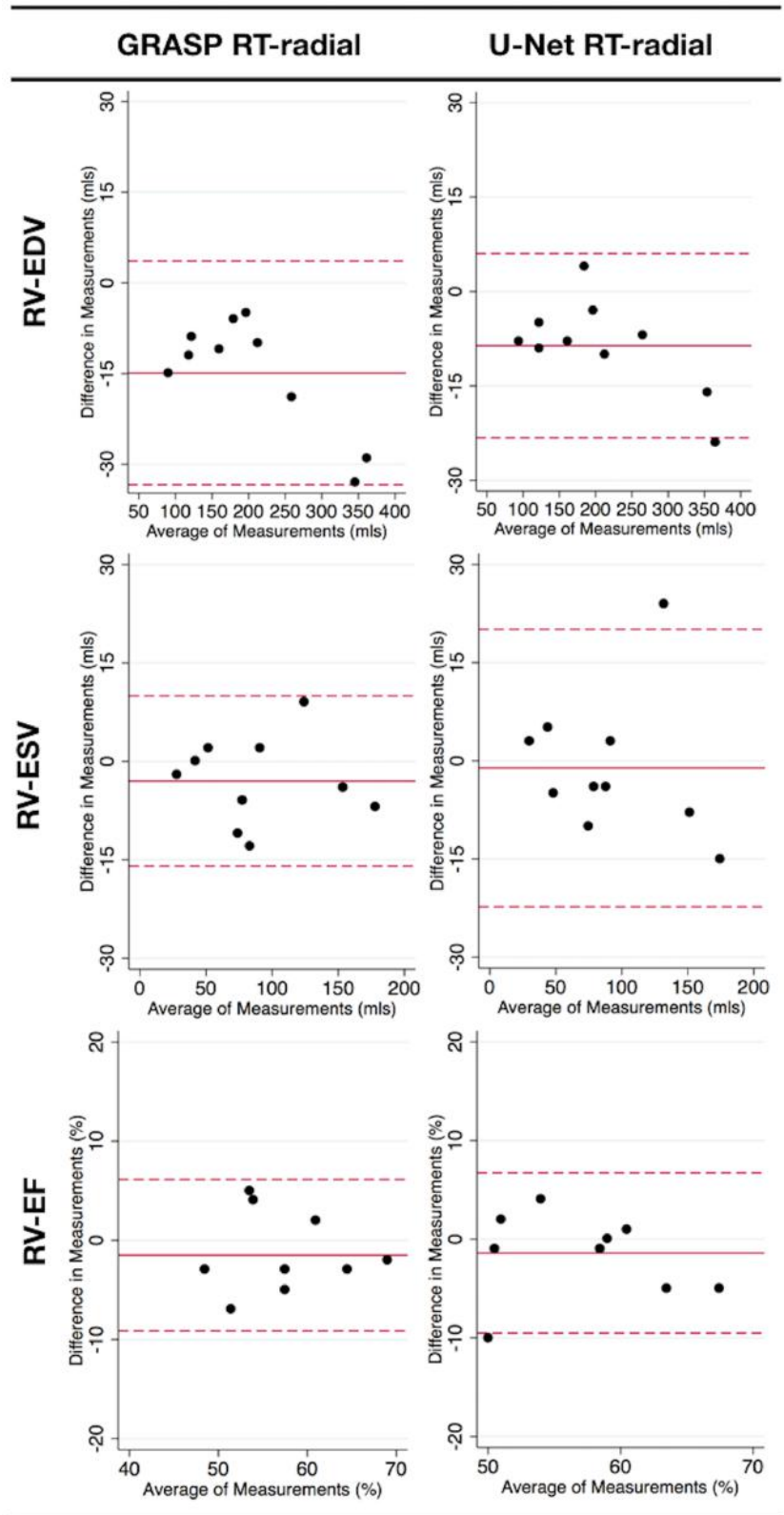

**Supporting Information Figure S13**

*Comparison of $\ell^2$-loss and $\ell^1$-loss functions on accuracy of reconstruction using synthetic test data. a) 'Truth' image from one subject, b) Reconstruction using network with $\ell^1$-loss function, c) Reconstruction using network with $\ell^2$-loss function, d) average RMSE and SSIM results over all synthetic data sets, from the networks trained with $\ell^2$-loss and $\ell^1$-loss functions, e) SSIM map of (b) from the result using the $\ell^1$-loss function, f) SSIM map of (c) from the result using the $\ell^2$-loss function.*

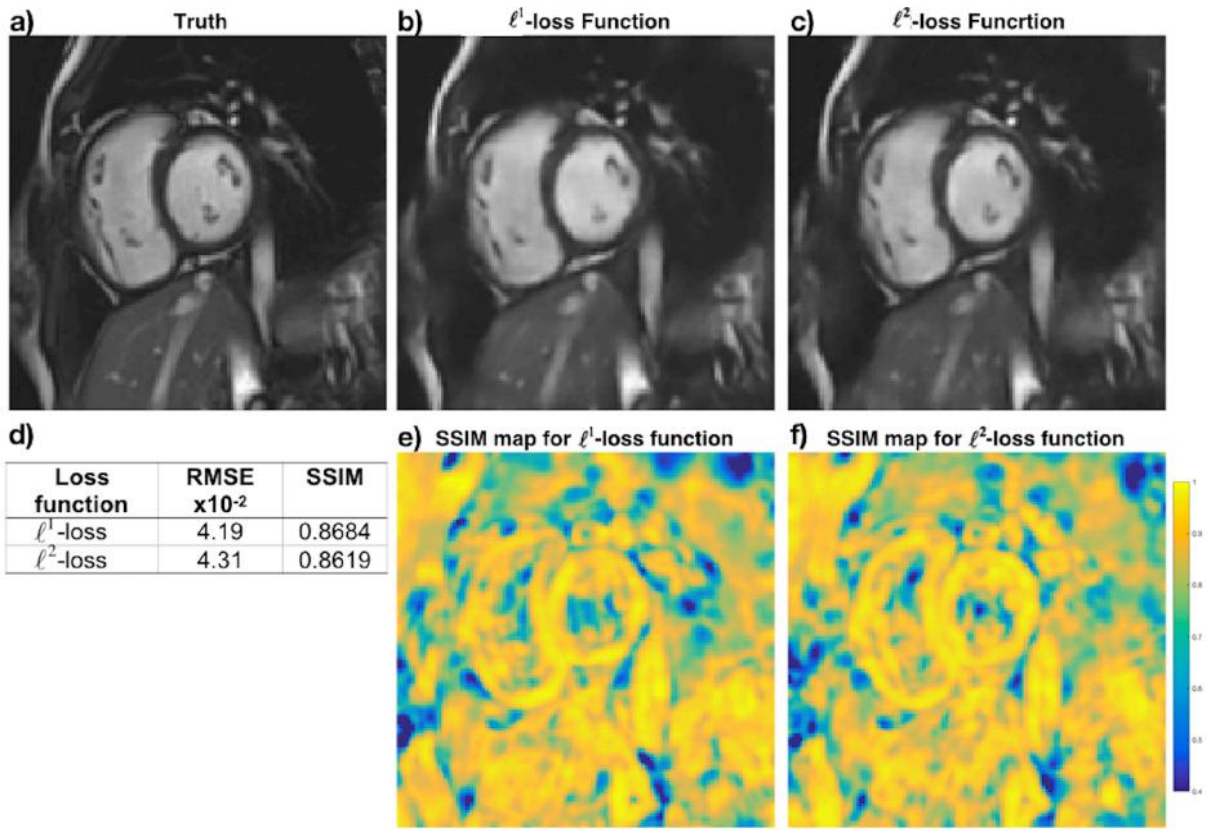